# Material Classification in the Wild: Do Synthesized Training Data Generalise Better than Real-World Training Data?

Grigorios Kalliatakis[1], Anca Sticlaru[1], George Stamatiadis[1], Shoaib Ehsan[1], Ales Leonardis[2], Juergen Gall[3] and Klaus D. McDonald-Maier[1]

[1]School of Computer Science and Electronic Engineering, University of Essex, Colchester, UK
[2]School of Computer Science, University of Birmingham, Birmingham, UK
[3]Institute of Computer Science, University of Bonn, Bonn, Germany
{gkallia, asticl, gstama, sehsan, kdm}@essex.ac.uk, a.leonardis@cs.bham.ac.uk, gall@iai.uni-bonn.de

Abstract: We question the dominant role of real-world training images in the field of material classification by investigating whether synthesized data can generalise more effectively than real-world data. Experimental results on three challenging real-world material databases show that the best performing pre-trained convolutional neural network (CNN) architectures can achieve up to 91.03% mean average precision when classifying materials in cross-dataset scenarios. We demonstrate that synthesized data achieve an improvement on mean average precision when used as training data and in conjunction with pre-trained CNN architectures, which spans from ~ 5% to ~ 19% across three widely used material databases of real-world images.

## 1 INTRODUCTION

Material classification in real-world environments is a challenging problem due to the huge impact of viewing and illumination conditions on material appearance. Therefore, training an appropriate classifier requires a training set which covers all these conditions as well as the intra-class variance of the materials. Two approaches are mainly used to generate suitable training sets: 1) capture a single representative per material category under a multitude of different conditions, such as scale, illumination and viewpoint, in a controlled setting [Caputo et al., 2005, Dana et al., 1999, Hayman et al., 2004, Liu et al., 2013] 2) use images acquired under uncontrolled conditions (e.g., FMD [Sharan et al., 2010] which is generated by taking images from an internet image database (Flickr)).

For (1), the measured viewing and illumination configurations are rather coarse and hence not descriptive enough to capture the mesoscopic effects in material appearance, which consider the light interaction with material surface regions mapped to approximately one pixel, in an accurate way. Moreover, the material samples are only measured under controlled illumination or lab environments which may not generalise to material appearance under complex realworld scenarios. On the other hand, approach (2) has the advantage that both the intra-class variance of materials and the environment conditions are sampled in a representative way. Unfortunately, the images have to be collected manually, and the materials appearing in the image have to be segmented and annotated. The necessary effort again severely limits the number of configurations that can be generated this way.

Motivated by the success of synthesized data for different vision applications (e.g. [Vazquez et al., 2014, Enzweiler and Gavrila, 2008, Pishchulin et al., 2011, Targhi et al., 2008, Stark et al., 2010, Shotton et al., 2011, Oxholm and Nishino, 2012, Barron and Malik, 2012, Barron and Malik, 2013]), we question the dominant role of real-world images in the field of material classification and investigate methodically whether synthetic datasets generalise better than real-world datasets when applied as training datasets and in conjunction with CNN architectures. For performing the large set of experiments, we partly followed the approach of Chatfield et al. [Chatfield et al., 2014] which was used for comparing CNN architectures for recognition of object categories. We, on the other hand, tackle material classification in this particular work, an entirely different problem from [Chatfield et al., 2014]. Continuing from our previous work [Kalliatakis et al., 2017], we go one step further by investigating the effect of training datasets (both real-world and synthetic) on the accuracy of the system. Our experimental results on three challenging real-world materials databases show that the best performing pre-trained CNN architectures can achieve up to 91.03% mean average precision for cross-dataset scenarios. We also demonstrate that synthesized data achieve an improvement on mean average precision when used as training data, which spans from ~ 5% to ~ 19% across three widely used materials databases of real-world images and in conjunction with pre-trained CNN architectures.

The rest of the paper is organised as follows. Section 2 gives details of the material classification pipeline used for our experiments. Section 3 investigates whether synthetic data generalise better than real-world images for the material classification task when applied as training datasets and in conjunction with pre-trained CNN architectures and presents experimental results in this regard. Finally, conclusions are given in Section 4.

## 2 MATERIAL CLASSIFICATION PIPELINE

Every block in the material classification pipeline is fixed except the feature extractor as different CNN architectures (pre-trained on 1000 ImageNet classes) are plugged in, one at a time, to compare their performance utilizing the mean average precision (mAP) metric. Given a training dataset $T_r$ consisting of m material categories, a test dataset $T_s$ comprising unseen images of the material categories given in $T_r$, and a set of n pre-trained CNN architectures ($C_1,...C_n$), the pipeline operates as follows: The training dataset $T_r$ is used as input to the first CNN architecture $C_1$. The output of $C_1$ is then utilized to train m SVM linear classifiers. Once trained, the test dataset $T_s$ is employed to assess the performance of the material classification pipeline using mAP. The training and testing procedures are then repeated after replacing $C_1$ with the second CNN architecture $C_2$ to evaluate the performance of the material classification pipeline. For a set of n pre-trained CNN architectures, the training and testing processes are repeated n times. Since the whole pipeline is fixed (including the training and test datasets, learning procedure and evaluation protocol) for all n CNN architectures, the differences in the performance of the material classification pipeline can be attributed to the specific CNN architectures used.

Following [Chatfield et al., 2014], we have chosen three baseline CNN architectures for our experiments, namely Fast (CNN-F), Medium (CNN-M) and Slow (CNN-S). The CNN-F architecture is similar to the one used by Krizhevsky et al. [Krizhevsky et al., 2012]. On the other hand, the CNN-M architecture is similar to the one employed by Zeiler and Fergus [Zeiler and Fergus, 2014], where as the CNN-S architecture is related to the 'accurate' network from the OverFeat package [Sermanet et al., 2013]. All these baseline CNN architectures are built on the Caffe framework [Jia et al., 2014] and are pre-trained on ImageNet [Deng et al., 2009]. Each network comprises 5 convolutional and 3 fully connected layers for a total of 8 learnable layers used to extract the material features from the images presented. Since transfer learning is applied, the CNN model output discussed above refers to the activations of the fc6 fully connected layer, which is the layer before the last. This is applied to all three CNN architectures used in the current pipeline. Data augmentation is used throughout the paper as it provides informative samples from current images for the system to use and increase the amount of training data. For instance, if we take the first row in in Table 1, it means that the C+F augmentation is used to generate the new images during the training phase and is further combined with the sum pooling in the testing phase (s). For further design and implementation details for these architectures, please see Table 1 in [15].

Table 1: Cross-dataset material classification results. Training and testing are performed using 3 different databases of real-world images. The name on the left denotes the training database, while the name on the right implies the testing database. Bold font highlights the leading mean result for every experiment. Three data augmentation strategies are used for both training and testing: 1) no augmentation (denoted Image Aug=-), 2) flip augmentation (denoted Image Aug=(F)), 3) crop and flip (denoted Image Aug=(C)). Augmented images are used as stand-alone samples (f), or by combining the corresponding descriptors using sum (s) or max (m) pooling or stacking (t). Here, GS denotes gray scale. The same symbols for data augmentation options and gray scale are used in the rest of the paper. For instance, if we take the first row, it means that the C+F augmentation is used to generate the new images during the training phase and is further combined with the sum pooling in the testing phase (s).

| Method | Image Aug. Training | Testing | FMD - ImageNet mAP | FMD - MINC-2500 mAP | MINC-2500 - ImageNet7 mAP |
|---|---|---|---|---|---|
| (m) CNN F | (C) f | s | 78.23% | 71.87% | 85.11% |
| (n) CNN S | (C) f | s | 83.49% | 72.95% | 86.18% |
| (o) CNN M | | | 82.40% | 73.06% | 87.64% |
| (p) CNN M | (C) f | s | 81.68% | 74.82% | 85.79% |
| (q) CNN M | (C) f | m | 81.69% | 75.46% | 86.55% |
| (r) CNN M | (C) s | s | 79.52% | 73.56% | 89.88% |
| (s) CNN M | (C) t | t | 80.22% | 74.19% | 89.53% |
| (t) CNN M | (C) f | | 80.31% | 73.83% | 82.71% |
| (u) CNN M | (F) f | | 81.91% | 73.01% | 91.03% |
| (v) CNN M GS | f | | 71.82% | 66.78% | 89.37% |
| (w) CNN M GS | (C) | s | 75.95% | 69.05% | 87.87% |
| (x) CNN M 2048 | (C) f | s | 80.27% | 76.35% | 86.82% |
| (y) CNN M 1024 | (C) f | s | 82.55% | 74.85% | 87.89% |
| (z) CNN M 128 | (C) f | s | 82.90% | 73.99% | 88.13% |

# 3 SYNTHETIC DATA VS REAL-WORLD DATA FOR TRAINING

For synthetic images, perfect segmentations are usually available without the need for manual segmentation, and a huge number of them can be generally obtained automatically. The system presented uses patch segments to train our material system to identify the correct feature points of a selected category to overcome the problems of material datasets that consist of small fraction of possible real-world conditions under a controlled laboratory environment. Recently, it was shown by [Weinmann et al., 2014] that real-world materials can be classified using synthesized training data. In this work, we challenge the dominant role of real-world images in the field of material classification and methodically analyze how synthetic data affects the results when applied as training sets and used in conjunction with pre-trained CNN architectures. We use cross-dataset analysis in this section to demonstrate how well the real-world and synthetic training datasets perform in complex real-world scenarios by employing real-world test datasets. The synthetic data is generated using a different database to the one used for training which will still fit within our system's aim: to train on one dataset and test on a different dataset.

Training m linear SVM classifiers, we take into consideration for our analysis only the classes that are common across both the synthetic and real-world data. For instance, in this type of scenario, if synthetic data is chosen for training and a real-world dataset such as FMD is chosen for testing, then only the four common classes such as fabric, leather, stone and wood, are selected to be included in the analy- sis. The same would be achieved for when real-world database is chosen for training and synthetic for testing.

## 3.1 Material Databases

Three different real-world databases are used in our experiments: 1) Flickr Material Database (FMD) [Sharan et al., 2010], 2) ImageNet7 dataset [Hu et al., 2011] which was derived from ImageNet [Deng et al., 2009] by collecting 7 common material categories, and 3) MINC-2500 [Bell et al., 2015] which is a patch classification dataset with 2500 samples per category. For synthetic data, University of Bonn dataset (UBO2014) [Weinmann et al., 2014] is used for training purposes and tested against FMD, ImageNet7 and MINC-2500 datasets. This synthetic dataset consists of 7 material categories with 12 different item samples per category. Each material sample is measured and consists of 151 diverse lighting directions and 151 viewing direction of the of it resulting in 22,801 images per category and 159,607 images in total. As evident, all three real-world databases consist of neither the same number of images nor categories between them. For this specific reason and in order to keep the tests on a common base, we consider the first half of the images enclosed in each database category as positive training samples and the other half for testing. Regarding negative training samples, for a category, the first 10% images of the remaining classes from the dataset are collected together. Generating the negative training subset this way means the category still has images from the same dataset, but they will be different from the evaluated category. Finally, a dataset containing 1414 random images is utilized and kept constant as the negative test data of our system for all the experiments that follow. In total, 14 different variants of the baseline CNN architectures with different data augmentation strategies are compared on FMD, ImageNet7 and MINC-2500.

Table 2: Material classification results using synthesized images. Training is performed using synthesized data from [Weinmann et al., 2014]. Bold font highlights the leading mean result obtained when tested on databases of real-world images (FMD, MINC-2500, ImageNet7).

| Method | | Image Aug. Training | Testing | MINC-2500 mAP | ImageNet7 mAP | FMD mAP |
|---|---|---|---|---|---|---|
| (a) CNN F | (C) | f | s | 94.99% | 95.41% | 78.76% |
| (b) CNN S | (C) | f | s | 94.56% | 95.61% | 78.23% |
| (c) CNN M | | | | 93.36% | 95.76% | 74.72% |
| (d) CNN M | (C) | f | s | 95.16% | 95.19% | 77.30% |
| (e) CNN M | (C) | f | m | 95.10% | 95.11% | 76.79% |
| (f) CNN M | (C) | s | s | 95.62% | 96.17% | 79.26% |
| (g) CNN M | (C) | t | t | 95.52% | 96.20% | 78.95% |
| (h) CNN M | (C) | f | | 94.24% | 94.06% | 70.91% |
| (i) CNN M | (F) | f | | 93.43% | 95.64% | 74.25% |
| (j) CNN M GS | | | | 91.58% | 93.32% | 63.17% |
| (k) CNN M GS | (C) | f | s | 94.77% | 93.48% | 70.34% |
| (l) CNN M 2048 | (C) | f | s | 95.42% | 95.06% | 77.01% |
| (m) CNN M 1024 | (C) | f | s | 94.89% | 94.81% | 69.84% |
| (n) CNN M 128 | (C) | f | s | 95.25% | 93.49% | 63.53% |

## 3.2 Cross-Dataset Analysis with Real-World Images

Results for three different cross-dataset experiments are given in Table 1: 1) Training on FMD and testing on ImageNet7 2) Training on FMD and testing on MINC-2500 3) Training on MINC-2500 and testing on ImageNet7. Considering the fact that FMD dataset is quite small, with only 100 images per ma-terial class, it performs better when used for training with reduced feature dimensionality per image, also observed in Zheng et al [Csurka et al., 2004]. In Table 1, with FMD as training database, the material classification pipeline performs best in testing the overlapping categories with ImageNet7 when Medium CNN architecture is used with 128 feature points per image extracted. The crop and flip augmentation and sum pooling collation is also used in this configuration and an accuracy of ~ 82% is achieved across the metric used. For FMD as training and MINC-2500 as testing database, the material classification pipeline achieves the best accuracy in testing the overlapping categories when CNN-M architecture is utilised with 2048 feature points per image extracted. Crop and flip augmentation and sum pooling are also used and the resulting accuracy is ~ 76% across the metric used. It is evident from Table 1 that the performance of the system increases when MINC-2500 is used as training database and overlapping categories of ImageNet7 are tested. This is due to the fact that MINC-2500 database enables the use of more images for positive training. When testing the overlapping categories with ImageNet7. In this case, the highest accuracy is again achieved when CNN-M is used. However, only flip is used as augmentation and no collation is utilised with this CNN architecture as opposed to the above two cases. The resulting accuracy of the system is ~ 90%. This is the case of finding the best balance before over-fitting occurs. Finally, the resulting average across all three experiments is ~ 82%.

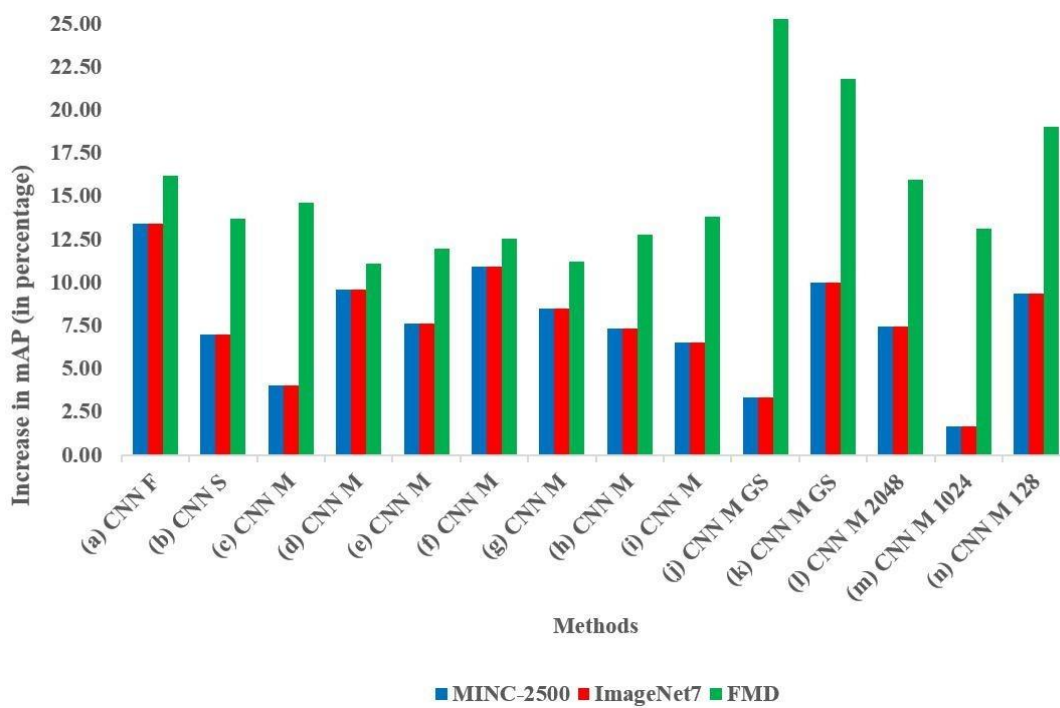

Figure 1: Increase in Mean average precision (mAP) of the material classification pipeline when UBO2014 database is used as training dataset and testing is done on MINC-2500, ImageNet7 and FMD as compared to the results obtained when training and testing datasets are both generated by using the same database of real-world images (e.g., training on FMD and testing on FMD).

### 3.3 Cross-Dataset Analysis with Synthesized Images

By using the UBO2014 database [Weinmann et al., 2014] for training, there is considerable increase in performance of the material classification pipeline as compared to when trained on databases with real-images as shown in Table 2. In the case of using UBO2014 for training and MINC-2500 for testing, there is a ~ 19% increase in performance, compared to when FMD was used for training, giving an accuracy of ~ 95%. Testing on ImageNet7 also gives UBO2014 the edge over FMD and MINC-2500. When comparing with FMD as test database, the increase in performance is quite significant at ~ 14%. This can be attributed to the fact that FMD is a small size real-world dataset that provides limited training and testing data, whereas UBO2014 covers a large fraction of viewing and illumination conditions on material appearance as well as the intra-class variance of the materials. This outlined reason for the significant increase in performance shows the importance of the amount of training data as well as the necessity of having bigger defined existing real-world datasets. When comparing UBO2014 training to MINC-2500 training, an increase in performance of ~ 5% is also observed which gives the synthetic data a better generalization in all tests and classes of materials for the datasets we used. Finally, Fig. 1 shows the increase in mean average precision (mAP) of the material classification pipeline when UBO2014 database is used as training dataset and testing is done on MINC-2500, ImageNet7 and FMD as compared to the results obtained when training and testing datasets are both generated by using the same database of real-world images (e.g., training on FMD and testing on FMD). It is thus clear from Fig. 1 that training with synthetic data can even surpass the performance that is achieved by training on the same database of real-world images on which it is tested.

## 4 CONCLUSIONS

We have done rigorous investigation of the effects of training data type for the task of material classification in the wild utilizing pre-trained CNN architectures. Our cross-dataset experiments have demonstrated that synthetic training data generalises better for material classification than real world training data. It seems that synthetic data is a promising approach to overcome the dataset bias problem of datasets collected from real images. It will be an interesting future direction to investigate if synthetic data can be combined with real images to improve accuracy and generalisation abilities of CNNs.